\documentclass{article}

\usepackage{spconf,amsmath,graphicx}
\usepackage[pagebackref=true,breaklinks=true,colorlinks,bookmarks=false]{hyperref}

\usepackage{times}
\usepackage{epsfig}
\usepackage{graphicx}
\usepackage{amsmath}
\usepackage{amssymb}
\usepackage{booktabs}

\title{Global-Local Propagation Network for RGB-D Semantic Segmentation}
\name{Sihan Chen\textsuperscript{\rm{1,2}}, Xinxin Zhu\textsuperscript{\rm{1}}, Wei Liu\textsuperscript{\rm{1,2}}, Xingjian He\textsuperscript{\rm{1,2}}, Jing Liu\textsuperscript{\rm{1,2}}}
\address{\textsuperscript{1} National Laboratory of Pattern Recognition, Institute of Automation, Chinese Academy of Sciences \\
\textsuperscript{2} School of Artificial Intelligence, University of Chinese Academy of Sciences}

\begin{document}
%\ninept
%
\maketitle

\begin{abstract}
   Depth information matters in RGB-D semantic segmentation task for providing additional geometric information to color images. Most existing methods exploit a multi-stage fusion strategy to propagate depth feature to the RGB branch. However, at the very deep stage, the propagation in a simple element-wise addition manner can not fully utilize the depth information. We propose Global-Local propagation network (GLPNet) to solve this problem. Specifically, a local context fusion module(L-CFM) is introduced to dynamically align both modalities before element-wise fusion, and a global context fusion module(G-CFM) is introduced to propagate the depth information to the RGB branch by jointly modeling the multi-modal global context features. Extensive experiments demonstrate the effectiveness and  complementarity of the proposed fusion modules. Embedding two fusion modules into a two-stream encoder-decoder structure, our GLPNet achieves new state-of-the-art performance on two challenging indoor scene segmentation datasets, i.e., NYU-Depth v2 and SUN-RGBD dataset. 
\end{abstract}
\begin{keywords}
rgb-d, sementic segmentation
\end{keywords}
\section{Introduction}
\label{sec:intro}

With the rapid development of RGB-D sensors like Microsoft Kinect, people get access to depth data more easily. Depth data can naturally describe 3D geometric information and reflect the structure of objects in the scene, thus can serve as a complementary modality for RGB data which captures rich color and texture information to improve semantic segmentation results. However, how to fully utilize depth information and effectively fuse the  two complementary modalities still remains an open problem.

Early approaches  \cite{long2015fully} attempt to use a two-stream network to extract features from RGB and depth modality respectively and fuse them at the last layer to predict the final segmentation results. This kind of ``late fusion" strategy fuses two modalities too late, resulting in that the RGB branch can not get needed geometric information guidance at the early stages. Later, researchers tend to propagate features of the depth branch to the RGB branch in a multi-stage manner, i.e., add depth feature to the RGB branch at the end of every stage of the encoder network  \cite{hazirbas2016fusenet,jiang2018rednet}. This strategy leverages geometric cues more early and sufficiently, which is proved to be effective. 

However, we rethink that the feature propagation way in conventional methods can not fully utilize the depth information due to the following two reasons. Firstly, the consecutive use of convolution and down-sampling operations in the two-stream encoder makes the two modality features are not aligned  each other as well as the early layers, so the geometric information provided by depth feature at the deep stage is not precise enough to assist RGB feature. Secondly, compared with the depth information, the RGB information reflects more semantic information, and the fact is  more obvious at the higher level of the network. Thus, the element-wise addition is not an appropriate solution, and the RGB should receive more focus for the semantic prediction.

\begin{figure*}
\begin{center}
\includegraphics[width=1.0\textwidth]{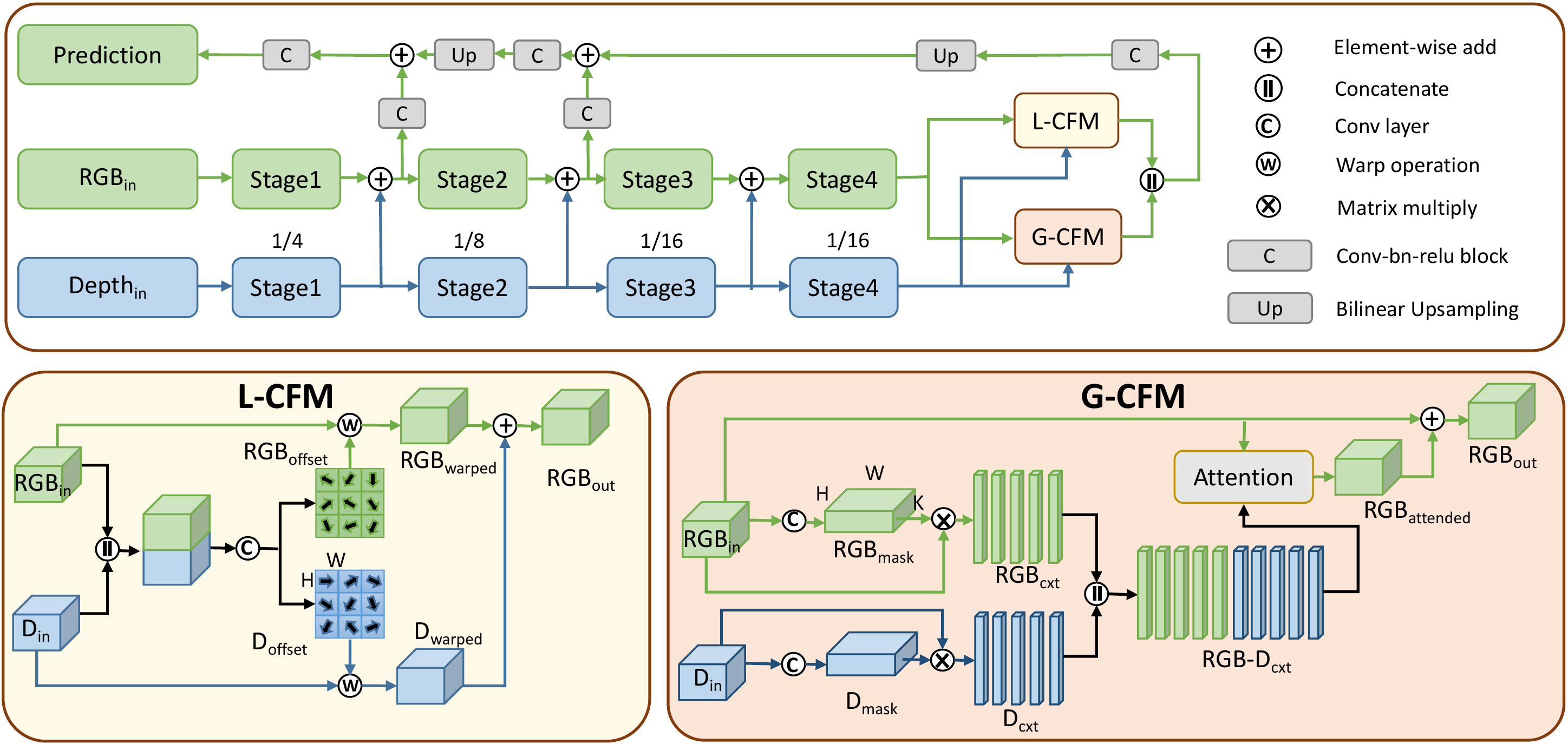}
\end{center}
\vspace{-0.7cm}
   \caption{An overview of our GLPNet. The fraction number describes the resolution ratio to the raw input image. We use dilation strategy in the last stage and the overall stride is 16. Best viewed in color.}
\vspace{-0.3cm}
\label{fig:overall-architecture}
\end{figure*}

To address the above problems, we propose Global-Local Propagation Network (GLPNet) to jointly utilize the complementary modalities of the depth and RGB features, in which a local context fusion module (L-CFM) and a global context fusion module (G-CFM) are designed to solve the problems of spatial disalignment and  semantic propagation in the feature fusion, respectively. Instead of directly adding depth feature to the RGB branch, L-CFM dynamically aligns the features of both modalities first before modality fusion. Specifically, the alignment process is to simultaneously warp the feature maps of both modalities according to the offsets predicted through a convolution layer which is inspired by the optical flow in video processing field and Semantic Flow \cite{li2020semantic}. Besides, the G-CFM is proposed to propagate the depth feature to the RGB branch through the joint multi-modal context modeling. Concretely, we extract global context features from both modalities and aggregate them to every RGB pixel using attention mechanism. Compared to L-CFM which precisely aligns the local features of two modalities, G-CFM aims at utilizing depth information from the view of global context. Given that the proposed two fusion modules help depth feature propagation from orthogonal perspectives (i.e., globally and locally),
combining them in a parallel way further improves the propagation effectiveness  at deep stage.

 The proposed GLPNet achieves new state-of-the-art performance on two challenging RGB-D semantic segmentation datasets, i.e., NYU-Depth v2   and SUN-RGBD dataset.

\section{Framework}
\label{sec:format}

\subsection{Overview}
The overall framework of our approach is depicted in Figure \ref{fig:overall-architecture} which uses an encoder-decoder architecture. In the encoder part, we use a two-stream backbone network (e.g. ResNet 101) to extract features from both modalities separately like the previous methods \cite{hazirbas2016fusenet,jiang2018rednet,park2017rdfnet,chen2020bi}. We take the multi-stage fusion strategy to propagate depth feature to the RGB branch. Specifically, we simply propagate depth feature in an element-wise addition manner for the three early stages and through the designed local and global fusion modules for the last stage. After applying those two fusion modules parallelly, we concatenate the outputs of them and add an additional convolution block to further process the fused feature then feed it into the segment decoder to get the final prediction. The segment decoder takes a FPN \cite{lin2017feature} -like structure which gradually upsamples the feature map and merges the shallow stage features ( i.e., stage-1, stage-2) through skip connections and the channel dimension of different stage features are reduced to 256 before fusion.

% \begin{figure}[t]
% \begin{center}
% %\fbox{\rule{0pt}{2in} \rule{0.9\linewidth}{0pt}}
% \includegraphics[width=1.0\linewidth]{L-CFM.png}
% \end{center}
%   \caption{The details of Local Context Fusion Module.}
% \label{fig:L-CFM}
% \end{figure}

\subsection{Local Context Fusion Module}
We introduce Local Context Fusion Module to dymanically adjust both modality features before their summation to help depth feature propagation. As illustrated at the bottom left corner of Figure \ref{fig:overall-architecture},  we take both modality features of the last stage as the inputs to the L-CFM, denoted as $\text {RGB} _{in} \in \mathbb{R}^{C\times H \times W} $ and $\text {D} _{in} \in  \mathbb{R}^{C\times H \times W} $,  respectively. From an intuitive perspective, the dynamic alignment should be inferred according to the spatial relations of both modalities, so we concatenate the two modality features along the channel dimension and then apply a convolution layer to predict the offset fields for each modality, denoted as $\text {RGB} _{offset} \in  \mathbb{R}^{2\times H \times W} $ and $\text {D} _{offset} \in  \mathbb{R}^{2\times H \times W} $, respectively. Then we use warp operation to adjust the features of both modalities according to the predicted offset fields separately and add the aligned depth feature to the aligned RGB feature to get the final output. More details about the warp operation can be found in the supplementary materials.

% \begin{figure*}[t]
% \begin{center}
% %\fbox{\rule{0pt}{2in} \rule{0.9\linewidth}{0pt}}
% \includegraphics[width=1.0\linewidth]{G-CFM.png}
% \end{center}
%   \caption{The details of Global Context Fusion Module.}
% \label{fig:G-CFM}
% \end{figure*}

\subsection{Global Context Fusion Module}
Inspired by the context researches on semantic segmentation which focuses on single RGB modality, we exploit the multi-modal global context information to further help depth feature propagation. The details of the G-CFM is illustrated at the bottom right corner of Figure \ref{fig:overall-architecture}. Just like the L-CFM, it also takes the RGB and depth feature of the last stage as input, i.e., $\text {RGB} _{in} \in  \mathbb{R}^{C\times H \times W} $ and $\text {D} _{in} \in  \mathbb{R}^{C\times H \times W} $ . We apply two independent convolution layers and softmax function along the spatial dimension to compute the pooling masks of both modalities, denoted as $\text {RGB} _{mask} \in  \mathbb{R}^{K\times H \times W} $ and $\text {D} _{mask} \in  \mathbb{R}^{K\times H \times W}$, respectively. $\text K$ is a hyperparameter which controls the number of global context feature vectors. We reshape the predicted  $ \text{RGB} _{mask}$ to $ \mathbb{R}^{K\times HW}$ and $ \text{RGB} _{in}$ to $\mathbb{R}^{HW\times C}$ and then perform a matrix multiplication to extract $\text K$ global context features $ \text{RGB} _{cxt} \in  \mathbb{R}^{K\times C}$. The same process is performed on the depth feature to compute $ \text{D} _{cxt} \in  \mathbb{R}^{K\times C}$ and then we concatenate two groups of context features to generate the multi-modal  context features $\text{RGB-D}_{cxt} \in \mathbb{R}^{2K\times C}$.

After modeling the  multi-modal global context features, we aggregate them back to the RGB feature using attention mechanism. Specifically, we feed $\text {RGB} _{in}$ into a $1\times 1$ convolution layer to generate query features $\text Q\in \mathbb{R}^{C^{'}\times H\times W}$ ($C^{'}=1/4C$), and multi-modal context features into two linear layers to generate the key features  $\text K\in \mathbb{R}^{C^{'}\times 2K}$ and value features  $\text V\in \mathbb{R}^{C\times 2K}$, respectively. Then we reshape $\text Q$ to $ \mathbb{R}^{C^{'}\times HW}$ and perform a matrix multiplication between the transpose of $\text Q$ and $\text K$ to calculate the attention map  $\text A\in \mathbb{R}^{HW\times 2K}$ and apply a softmax function to normalize the contribution of the 2k multi-modal context features. Finally, we multiply $\text A$ to the transpose of $\text V$ to compute the attended feature,  reshape and element-wisely add it to the original $\text{RGB}_{in}$ in a residual connection manner getting $\text{RGB}_{out}$ .
% \begin{equation}
%     \text A = \text{Q}^{T} \times \text K
% \end{equation}
% \begin{equation}
%     \text{RGB}_{out} = \text{RGB}_{in} + \text A \times \text V^{T} 
% \end{equation}

\section{Experiments}
\label{sec:pagestyle}

\subsection{Datasets and Implementation details}
To evaluate the proposed network, we conduct experiments on two RGB-D semantic segmentation datasets: NYU-Depth v2 \cite{silberman2012indoor} and SUN-RGBD \cite{song2015sun} dataset. 
NYU-Depth v2 dataset contains 1449 RGB-D images which are divided into 795 training images and 654 testing images. SUN-RGBD dataset consists of 10355 RGB-D images  which are divided into 5285 training images and 5050 testing images.

We choose two-stream dilated ResNet101 pretrained on ImageNet as the backbone network and the overall stride is 16. We use the SGD optimizer and employ a poly learning rate schedule. The initaial learning rate is set to 0.005 for NYU-Depth v2 dataset and 0.001 for SUN-RGBD dataset. Momentum and weight decay are set to 0.9 and 0.0005, respectively. Batch size is 8 for both datasets. We train the network by 500 epochs for the NYU-Depth v2 dataset and 200 epochs for the SUN-RGBD dataset. For data augmentation, we apply random scaling between $\left[ 0.5,2.25 \right]$, random cropping with crop size $480 \times 640$ and random horizontal flipping. We use cross-entropy as loss function. When assembling the segment decoder, we adopt the multi-loss strategy. Specifically, we compute two auxiliary losses using the stage-2 and stage-4 outputs of the decoder and the weight is both set to 0.2.
We report three metrics including pixel accuracy(Acc), mean accuracy(mACC) and mean intersection over union (mIoU).
\subsection{Ablation study}

We conduct extensive ablation experiments on the NYU-Depth v2 dataset to verify the effectiveness of the proposed modules. For the baseline model, we use conventional multi-stage propagation strategy that directly adds depth feature to the rgb branch at all four stages without using the decoder. As shown in Table \ref{tab:table_whole}, compared to the baseline model, the model embedded with L-CFM achieves a result of 48.22\% mIoU, which brings 1.49\% improvements and the model embedded with G-CFM achieves a result of 50.31\% mIoU, which brings 3.58\% improvements. When the L-CFM and G-CFM are used together in a parallel way, the result further improves to 51.39\% mIoU, demonstrating the proposed two fusion modules which propagate depth feature in  the local and global manner respectively are complement to each other. After assembling the segment decoder, we achieve 52.11\% mIoU. For fair comparison with the state-of-the-art, we adopt multi-grid and multi-scale test strategy and achieve 54.61\% mIoU. 
\begin{table}[ tbp]
    \centering
    \begin{tabular}{l|c}
    \toprule
        Method & mIoU\% \\
    \midrule
        baseline  & 46.73 \\
        +L-CFM & 48.22 \\
        +G-CFM & 50.31 \\
        +L-CFM +G-CFM &  51.39\\
        +L-CFM +G-CFM +decoder &52.11 \\
        +L-CFM +G-CFM +decoder +MG &53.57 \\
        +L-CFM +G-CFM +decoder +MG +MS&\textbf{54.61} \\
    \bottomrule
        
    \end{tabular}
    \caption{Ablation study  of proposed GLPNet on NYU-Depth v2 test set. MG: multi-grid. MS:multi-scale test. }
    \vspace{-0.5cm}
    \label{tab:table_whole}
\end{table}

 For the Local Context Fusion Module, we further conduct experiments with different settings of the locations where the L-CFM is embedded. As shown in Table \ref{tab:table_L-CFM}, using L-CFM at early stages can hardly improve the network performance given that the the alignment error has not been accumulated too big. We  further prove this by adding L-CFM to all stages and achieved a slight performance drop, demonstrating our design  that only adds it at the last stage can effectively perform precise depth feature propagation and overcome the accumulated alignment error.

% \begin{table}[ htbp]
%     \centering
%     \begin{tabular}{c|c|c|c|c}
%     \toprule
%          &S1 & S2 &S3 &S4\\
%     \midrule
%         RGB offset & 0.97&0.29 &0.41 &5.10  \\
%         D offset & 0.95& 0.99  &1.07 & 7.37  \\
        
%     \bottomrule
        
%     \end{tabular}
%     \caption{Mean values of the predicted offset vector norms on test set. The unit is pixel. S is short for Stage.}
%     \label{tab:offset_statistic}
% \end{table}

\begin{figure}[tbhp]
\begin{center}
\includegraphics[width=1.0\linewidth]{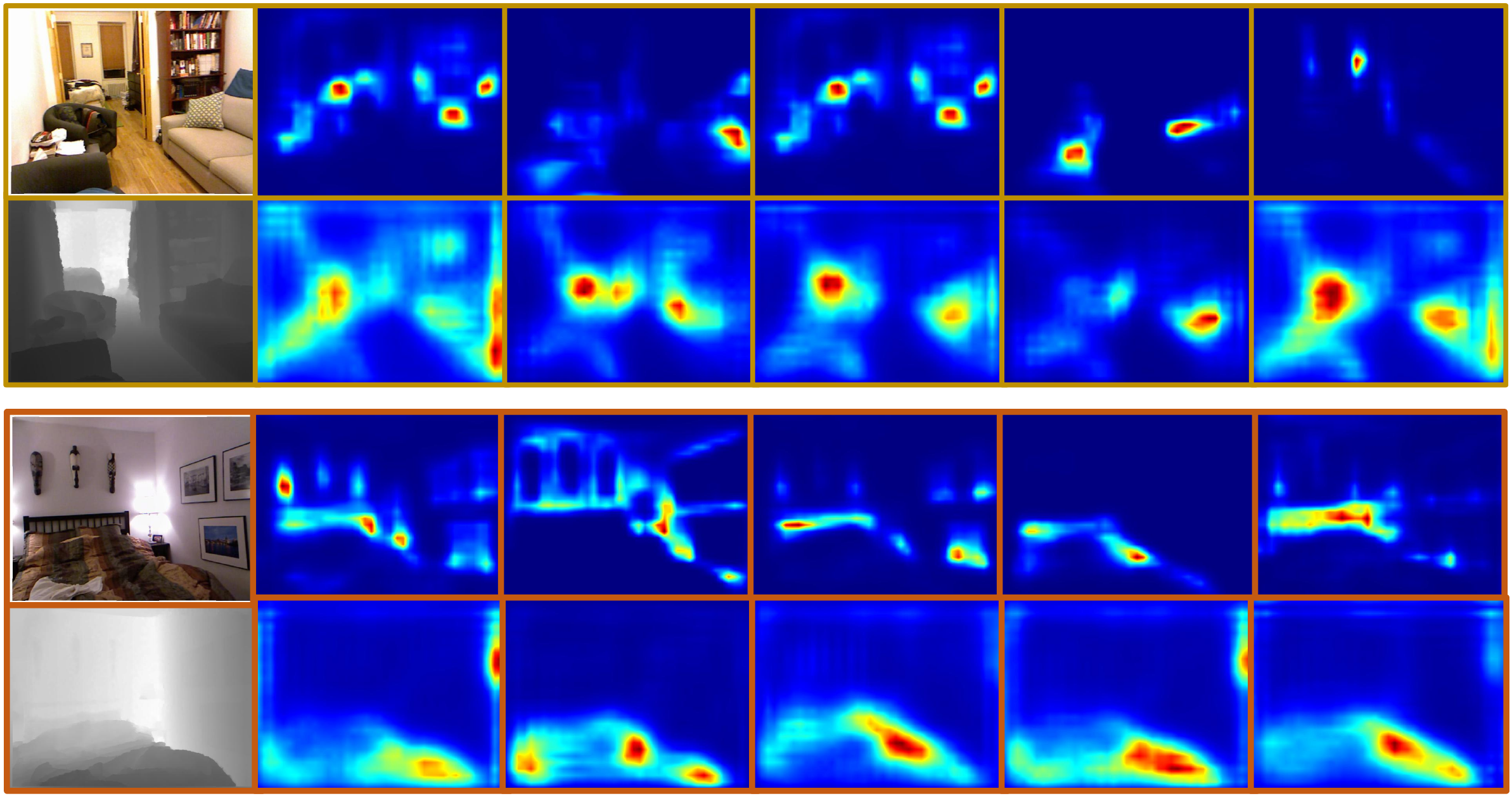}
\end{center}
  \vspace{-0.5cm}
  \caption{Visualization of the pooling masks predicted by the G-CFM for two example RGB-D pairs from NYU-Depth v2 test set. We  present five highly representative masks of fifteen for each modality to save space. Best viewed in color.}
  \vspace{-0.5cm}
\label{fig:mask_vis}
\end{figure}

For the Global Context Fusion Module, we further conduct two comparison experiments with different context modeling settings  denoted as G-CFM\_var1 and G-CFM\_var2 respectively. In G-CFM\_var1, we only extract K global context features from the input RGB feature and totally do not use the depth feature. In G-CFM\_var2, we directly add depth feature to the RGB branch like the baseline network and then extract K global context features from the fused feature.  The result is represented in Table \ref{tab:ablation-G_CFM}. We can see that G-CFM\_var1 which only models the RGB context yields 49.81\% mIoU.  G-CFM\_var2 can not properly utilize the depth information for that the unaligned depth feature brings noise and the performance drops to 49.56\% mIoU. Compared to those two variants, our G-CFM which exploits the multi-model context achieves 50.31 mIoU\% demonstrating the effectiveness and necessity to jointly model the multi-model contexts. With regards to the hyperparameter K , i.e., the number of global context features extracted by each modality, we found the network performance is sensitive to the its choices and we choose k=15 for its best performance. 
We choose two RGB-D image pairs from NYU-Depth v2 test set to visualize the pooling masks predicted by the G-CFM. The result is shown in  Figure \ref{fig:mask_vis} from which we can find that different global pooling masks concentrate on different regions of the images and are different between two modalities.

\begin{table}[ htbp]
    \centering
    \begin{tabular}{c c c c  c |c}
    \toprule
        Method &S1 & S2 &S3 &S4&mIoU\% \\

    \midrule
         & \checkmark& & &  &46.68\\
         & &\checkmark & &  &46.59 \\
        L-CFM & & &\checkmark &  &46.73 \\
         & & & & \checkmark & \textbf{48.22}\\
         & \checkmark&\checkmark &\checkmark & \checkmark &47.92 \\
    \bottomrule
        
    \end{tabular}
    \caption{Ablation study of different stages where L-CFM is embedded on NYU-Depth v2 test set. S represents stage.}
    \label{tab:table_L-CFM}
    \vspace{-0.5cm}
\end{table}

\begin{table}[ htbp]
    \centering
    \setlength{\tabcolsep}{8mm}
    \scalebox{0.8}{
    \begin{tabular}{l c|c}
    \toprule
        Method &K& mIoU\% \\
    \midrule
        G-CFM\_var1&15 & 49.81\\
        G-CFM\_var2&15  &49.56 \\
        G-CFM &15 &\textbf{50.31} \\
    \midrule
        G-CFM & 5 & 48.23\\
        G-CFM & 10 &49.27 \\
        G-CFM & 20 &50.27 \\
        G-CFM & 25 &49.42 \\
    \bottomrule
        
    \end{tabular}}
    \caption{Ablation study of different variants of G-CFM and the value of hyperparameter K on NYU-Depth v2 test set.}
    \vspace{-0.5cm}
    \label{tab:ablation-G_CFM}
\end{table}

\subsection{Comparison with the state-of-the-art}

% \begin{figure*}[htbp]
% \begin{center}
% %\fbox{\rule{0pt}{2in} \rule{0.9\linewidth}{0pt}}
% \includegraphics[width=1.0\linewidth]{visualization.png}
% \end{center}
%   \caption{Qualitative results on NYU-Depth v2 dataset.}
% \label{fig:visualization}
% \end{figure*}

\begin{table}[t]
    \centering
    \scalebox{0.8}{
    \begin{tabular}{c c c |c c c}
    \toprule
        Method & Backbone &DE& Acc\% & mAcc\% & mIoU \% \\
    \midrule
        FCN \cite{long2015fully}   &2$\times$VGG16 &HHA &65.4 & 46.1 & 34.0 \\
        LSD-GF \cite{cheng2017locality}   &2$\times$VGG16 &HHA &71.9 & 60.7 & 45.9 \\
        
        RDF101 \cite{park2017rdfnet}   &2$\times$Res101&HHA & 75.6 & 62.2 & 49.1 \\
        RDF152 \cite{park2017rdfnet}  &2$\times$Res152&HHA & 76.0 & 62.8 & 50.1 \\
        CFNet \cite{lin2017cascaded}  &2$\times$Res152&HHA & - & - & 47.7 \\
        3DGNN \cite{qi20173d}   &1$\times$VGG16 &HHA &- & 55.2 & 42.0 \\
        D-CNN \cite{wang2018depth}  &2$\times$VGG16&HHA & - & 56.3 & 43.9 \\
        ACNet \cite{hu2019acnet}    &3$\times$Res50&raw & - & - & 48.3 \\
        PADNet \cite{xu2018pad}   &1$\times$Res50 &raw & 75.2 & 62.3 & 50.2 \\
        PAP \cite{zhang2019pattern}   &1$\times$Res50 &raw & 76.2 & 62.5 & 50.4 \\
        SGNet-16s \cite{chen2020spatial}  &1$\times$Res101&raw & 76.4 & 62.1 & 50.3 \\
        SGNet-8s \cite{chen2020spatial}   &1$\times$Res101&raw & 76.8 & 63.1 & 51.0 \\
        % VCD+ACNet \cite{xiong2020variational}    &3$\times$Res50&raw &  - & 64.4 & 51.9 \\
        SA-Gate \cite{chen2020bi}   &2$\times$Res101 &HHA& 77.9 & - & 52.4
        \\
        
    \midrule
        Ours   &2$\times$ Res101 &raw& \textbf{79.1}& \textbf{66.6} & \textbf{54.6} \\
    \bottomrule
        
    \end{tabular}}
    \caption{Comparison results on NYU-Depth v2 test set. DE represents depth encoding. }
    
    \label{tab:table_NYUDv2}
\end{table}

\begin{table}[ htbp]
    \centering
    \scalebox{0.8}{
    \begin{tabular}{c c c |c c c}
    \toprule
        Method &  Backbone&DE & Acc\% & mAcc\% & mIoU \% \\
    \midrule
        LSD-GF \cite{cheng2017locality}   &2$\times$VGG16 &HHA &- & 58.0 & - \\
        3DGNN \cite{qi20173d}   &1$\times$VGG16 &HHA &- & 57.0 & 45.9 \\
        RDF152 \cite{park2017rdfnet}  & 2$\times$Res152&HHA &81.5 & 60.1 & 47.7 \\
        CFNet \cite{lin2017cascaded}  & 2$\times$Res152&HHA &- & - & 48.1 \\
        D-CNN \cite{wang2018depth}  &2$\times$VGG16&HHA & - & 53.5 & 42.0 \\
        ACNet \cite{hu2019acnet}   & 3$\times$Res50&raw& - & - & 48.1 \\
 
        PAP \cite{zhang2019pattern}   & 1$\times$Res50&raw &83.8 & 58.4 & 50.5 \\
        SGNet-8s \cite{chen2020spatial}   & 1$\times$Res101&raw &81.8 & 60.9 & 48.5 \\
        SA-Gate \cite{chen2020bi}   & 2$\times$Res101&HHA &82.5 & - & 49.4
        \\
    \midrule
        baseline  & 2$\times$Res101&raw & 79.3& 57.9 & 44.0     \\
        Ours   & 2$\times$Res101&raw&\textbf{82.8}& \textbf{63.3} &\textbf {51.2} \\
    \bottomrule
        
    \end{tabular}}
    \caption{Comparison results on SUN-RGBD test set.}
    \vspace{-0.5cm}
    \label{tab:table_SUN}
\end{table}

The performance comparison results on NYU-Depth v2 dataset is shown in Table \ref{tab:table_NYUDv2} and our method outperforms existing approaches with dominant advantage which demonstrates the effectiveness of our GLPNet with the proposed fusion modules. It is worth noting that we use raw depth image as the input to the depth branch and achieves better performance than those methods \cite{park2017rdfnet,lin2017cascaded,chen2020bi,wang2018depth} that take the encoded HHA \cite{gupta2014learning} images as input which consume much more inference time. Compared with the previous state-of-the-art \cite{chen2020bi} which uses a bi-direction information propagation strategy and uses deeplabv3+ \cite{chen2018encoder} to model the fusion context, our method surpass them over 2.2\% mIoU. we attribute it to the better depth feature propagation ability of our network with G-CFM modeling the multi-modal context information globally and L-CFM  
performing precise modality alignment locally.

% \subsubsection{Visualization}

% We visualize some examples on the test set of NYU-Depth v2 dataset and the qualitative result is shown in Figure \ref{fig:visualization} from which we can find that the proposed GLPNet can consistently improve the color image segmentation in different situations. Specifically, when depth changes continuously while the color changes drastically as shown in the first example and when depth is discriminative  while color is extremely similar between adjacent objects as shown in the second example, our method successfully utilize the depth information to improve results while the RGB-D baseline fails. In the last example, we show the case when depth information acts like noise and smooths the correct RGB prediction, this problem can also be solved by our method which avoids direct modality fusion in the deep stage.

We also conduct experiments on the SUN-RGBD dataset to further evaluate the proposed method. Quantitative results of this dataset are shown in Table \ref{tab:table_SUN}, our method boosts the RGB-D baseline from 44.0 mIoU\% to {51.2} mIoU\% and achieves the state-of-the-art.

% The VCD method propose by Xiong at al. \cite{xiong2020variational} can improves existing two-stream networks through alternating traditional convolution layers in the encoder to deformable convolution layers and sampling the weight according to the predicted gaussian kernel distribution. We assume this method is orthogonal to us and theoretically can boost our model samely (by alternating the convolution in our encoder to theirs). However, this is beyond our focus.

\section{Conclusion}
\label{sec:typestyle}

We have proposed GLPNet for RGB-D semantic segmentation. The GLPNet help the information propagation from the depth branch to RGB branch at deep stage. Specifically, the local context fusion module dynamically aligns both modalities before fusion and the global context fusion module performs depth information propagation through the joint multi-modal context modeling. Extensive ablation experiments have been conducted to verify the effectiveness of proposed method and the GLPNet achieved new  state-of-the-art performance on two indoor scene segmentation datasets, i.e., NYU-Depth v2 and SUN-RGBD dataset.

% \section{REFERENCES}
% \label{sec:ref}

\bibliographystyle{IEEEbib}
%\bibliography{strings,refs}
\bibliography{refs}

\begin{thebibliography}{10}

\bibitem{long2015fully}
Jonathan Long, Evan Shelhamer, and Trevor Darrell,
\newblock ``Fully convolutional networks for semantic segmentation,''
\newblock in {\em Proceedings of the IEEE conference on computer vision and
  pattern recognition}, 2015, pp. 3431--3440.

\bibitem{hazirbas2016fusenet}
Caner Hazirbas, Lingni Ma, Csaba Domokos, and Daniel Cremers,
\newblock ``Fusenet: Incorporating depth into semantic segmentation via
  fusion-based cnn architecture,''
\newblock in {\em Asian conference on computer vision}. Springer, 2016, pp.
  213--228.

\bibitem{jiang2018rednet}
Jindong Jiang, Lunan Zheng, Fei Luo, and Zhijun Zhang,
\newblock ``Rednet: Residual encoder-decoder network for indoor rgb-d semantic
  segmentation,''
\newblock {\em arXiv preprint arXiv:1806.01054}, 2018.

\bibitem{li2020semantic}
Xiangtai Li, Ansheng You, Zhen Zhu, Houlong Zhao, Maoke Yang, Kuiyuan Yang, and
  Yunhai Tong,
\newblock ``Semantic flow for fast and accurate scene parsing,''
\newblock {\em arXiv preprint arXiv:2002.10120}, 2020.

\bibitem{park2017rdfnet}
Seong-Jin Park, Ki-Sang Hong, and Seungyong Lee,
\newblock ``Rdfnet: Rgb-d multi-level residual feature fusion for indoor
  semantic segmentation,''
\newblock in {\em Proceedings of the IEEE international conference on computer
  vision}, 2017, pp. 4980--4989.

\bibitem{chen2020bi}
Xiaokang Chen, Kwan-Yee Lin, Jingbo Wang, Wayne Wu, Chen Qian, Hongsheng Li,
  and Gang Zeng,
\newblock ``Bi-directional cross-modality feature propagation with
  separation-and-aggregation gate for rgb-d semantic segmentation,''
\newblock {\em arXiv preprint arXiv:2007.09183}, 2020.

\bibitem{lin2017feature}
Tsung-Yi Lin, Piotr Doll{\'a}r, Ross Girshick, Kaiming He, Bharath Hariharan,
  and Serge Belongie,
\newblock ``Feature pyramid networks for object detection,''
\newblock in {\em Proceedings of the IEEE conference on computer vision and
  pattern recognition}, 2017, pp. 2117--2125.

\bibitem{silberman2012indoor}
Nathan Silberman, Derek Hoiem, Pushmeet Kohli, and Rob Fergus,
\newblock ``Indoor segmentation and support inference from rgbd images,''
\newblock in {\em European conference on computer vision}. Springer, 2012, pp.
  746--760.

\bibitem{song2015sun}
Shuran Song, Samuel~P Lichtenberg, and Jianxiong Xiao,
\newblock ``Sun rgb-d: A rgb-d scene understanding benchmark suite,''
\newblock in {\em Proceedings of the IEEE conference on computer vision and
  pattern recognition}, 2015, pp. 567--576.

\bibitem{cheng2017locality}
Yanhua Cheng, Rui Cai, Zhiwei Li, Xin Zhao, and Kaiqi Huang,
\newblock ``Locality-sensitive deconvolution networks with gated fusion for
  rgb-d indoor semantic segmentation,''
\newblock in {\em Proceedings of the IEEE conference on computer vision and
  pattern recognition}, 2017, pp. 3029--3037.

\bibitem{lin2017cascaded}
Di~Lin, Guangyong Chen, Daniel Cohen-Or, Pheng-Ann Heng, and Hui Huang,
\newblock ``Cascaded feature network for semantic segmentation of rgb-d
  images,''
\newblock in {\em Proceedings of the IEEE International Conference on Computer
  Vision}, 2017, pp. 1311--1319.

\bibitem{qi20173d}
Xiaojuan Qi, Renjie Liao, Jiaya Jia, Sanja Fidler, and Raquel Urtasun,
\newblock ``3d graph neural networks for rgbd semantic segmentation,''
\newblock in {\em Proceedings of the IEEE International Conference on Computer
  Vision}, 2017, pp. 5199--5208.

\bibitem{wang2018depth}
Weiyue Wang and Ulrich Neumann,
\newblock ``Depth-aware cnn for rgb-d segmentation,''
\newblock in {\em Proceedings of the European Conference on Computer Vision
  (ECCV)}, 2018, pp. 135--150.

\bibitem{hu2019acnet}
Xinxin Hu, Kailun Yang, Lei Fei, and Kaiwei Wang,
\newblock ``Acnet: Attention based network to exploit complementary features
  for rgbd semantic segmentation,''
\newblock in {\em 2019 IEEE International Conference on Image Processing
  (ICIP)}. IEEE, 2019, pp. 1440--1444.

\bibitem{xu2018pad}
Dan Xu, Wanli Ouyang, Xiaogang Wang, and Nicu Sebe,
\newblock ``Pad-net: Multi-tasks guided prediction-and-distillation network for
  simultaneous depth estimation and scene parsing,''
\newblock in {\em Proceedings of the IEEE Conference on Computer Vision and
  Pattern Recognition}, 2018, pp. 675--684.

\bibitem{zhang2019pattern}
Zhenyu Zhang, Zhen Cui, Chunyan Xu, Yan Yan, Nicu Sebe, and Jian Yang,
\newblock ``Pattern-affinitive propagation across depth, surface normal and
  semantic segmentation,''
\newblock in {\em Proceedings of the IEEE Conference on Computer Vision and
  Pattern Recognition}, 2019, pp. 4106--4115.

\bibitem{chen2020spatial}
Lin-Zhuo Chen, Zheng Lin, Ziqin Wang, Yong-Liang Yang, and Ming-Ming Cheng,
\newblock ``Spatial information guided convolution for real-time rgbd semantic
  segmentation,''
\newblock {\em arXiv preprint arXiv:2004.04534}, 2020.

\bibitem{gupta2014learning}
Saurabh Gupta, Ross Girshick, Pablo Arbel{\'a}ez, and Jitendra Malik,
\newblock ``Learning rich features from rgb-d images for object detection and
  segmentation,''
\newblock in {\em European conference on computer vision}. Springer, 2014, pp.
  345--360.

\bibitem{chen2018encoder}
Liang-Chieh Chen, Yukun Zhu, George Papandreou, Florian Schroff, and Hartwig
  Adam,
\newblock ``Encoder-decoder with atrous separable convolution for semantic
  image segmentation,''
\newblock in {\em Proceedings of the European conference on computer vision
  (ECCV)}, 2018, pp. 801--818.

\end{thebibliography}

\end{document}